# Knowledge Base Approach for 3D Objects Detection in Point Clouds Using 3D Processing and Specialists Knowledge

Helmi Ben Hmida, Frank Boochs
Institut i3mainz, am Fachbereich Geoinformatik und
Vermessung, Fachhochschule Mainz, Lucy-Hillebrand-
Str. 255128 Mainz, Germany
e-mail: {helmi.benhmida, boochs}@geoinform.fh-mainz.de

Christophe Cruz, Christophe Nicolle
Laboratoire Le2i, UFR Sciences et Techniques
Université de Bourgogne
B.P. 47870, 21078 Dijon Cedex, France
e-mail: {christophe.cruz, cnicolle}@u-bourgogne.fr

*Abstract*—This paper presents a knowledge-based detection of objects approach using the OWL ontology language, the Semantic Web Rule Language, and 3D processing built-ins aiming at combining geometrical analysis of 3D point clouds and specialist's knowledge. Here, we share our experience regarding the creation of 3D semantic facility model out of unorganized 3D point clouds. Thus, a knowledge-based detection approach of objects using the OWL ontology language is presented. This knowledge is used to define SWRL detection rules. In addition, the combination of 3D processing built-ins and topological Built-Ins in SWRL rules allows a more flexible and intelligent detection, and the annotation of objects contained in 3D point clouds. The created WiDOP prototype takes a set of 3D point clouds as input, and produces as output a populated ontology corresponding to an indexed scene visualized within VRML language. The context of the study is the detection of railway objects materialized within the Deutsche Bahn scene such as signals, technical cupboards, electric poles, etc. Thus, the resulting enriched and populated ontology, that contains the annotations of objects in the point clouds, is used to feed a GIS system or an IFC file for architecture purposes.

I. INTRODUCTION

Surveying with 3D scanners is spreading all domains. With every new scanner model on the market, the instruments become faster, more accurate and can scan objects at longer distances [1]. Such a technology presents a powerful tool for many applications and has partially replaced traditional surveying methods since it can speed up field work significantly. This method allows the creation of 3D point clouds from objects or landscapes.

From the other side, the technical survey of facility aims to build a digital model based on geometric analysis. Such a process becomes more and more tedious. Especially, with the new terrestrial laser scanners, where a huge amount of 3D point clouds are generated. Within such a scenario, new challenges have seen the light where the basic one is to make the reconstruction process automatic and more accurate. Thus, early works on 3D point clouds have investigated the reconstruction and the recognition of geometrical shapes [2] [3] to resolve this challenge. In fact, such a problematic was investigated as a topic of the computer graphic and the signal processing research, where most works focused on segmentation or visualization aspects. As most-recent works, the new tendency related to the use of semantic has been explored [4]. As a main operation, the technical survey relies fundamentally on the object reconstruction process where considerable effort has already been invested to reduce the impact of time consuming, manual activities and to substitute them by numerical algorithms.

Unfortunately, most of algorithmic conceptions are data-driven and concentrate on specific features of the objects, being accessible to numerical models. By these models, which normally describe the behavior of geometrical (flatness, roughness, for example) or physical features (color, texture), the data are classified and analyzed. Basically, such strategies are static and not to allow a dynamic adjustment to the object or initial processing results. In further scenarios, an algorithm will be applied to the data producing better or minor results, depending on several parameters like image or point cloud quality, the completeness of object representation, the viewpoints position, the complexity of object features, the use of control parameters and so on. Consequently, there is no feedback to the algorithmic part in order to choose a different algorithm or reuse the same algorithm with changed parameters. This interaction is mainly up to the user who has to decide by himself, which algorithms to apply for which kind of objects and data sets. Often good results can only be achieved by iterative processing controlled by a human interaction.

These problems can be solved when further supplementary and guiding information is integrated into the algorithmic process chain for object detection and recognition, allowing to support the process of validation. Such an information might be derived from the context of the object itself and its behavior with respect to the data and/or other objects or from a systematic characterization of the parameterization and effectiveness of the algorithms to be used. As programming languages used in the context of numerical treatments are not dedicated to process knowledge, their condition of use is not flexible and makes the integration of semantic aspects difficult.

Ontologies are used to represent formally the knowledge of a domain. The basic ideas were to present knowledge using graphs and logical structure to make computers able to understand and process knowledge [5]. As most recent works, the tendency related to the use of semantic has been explored [4][10][21]. In fact, the assumption that knowledge will help the improvement of the automation, the accuracy and the result quality is shared by specialists of the point cloud processing. However, many questions remain without answers. How the detection process can get support within different knowledge about the scene objects and what is the impact of this knowledge compared to classic approach. In such scenario, knowledge about such objects has to include detailed information about the objects' geometry, structure, 3D algorithms, etc.

The technical survey of facilities, as a long and costly process, aims at building a digital model based on geometric analysis since the modeling of a facility as a set of vectors is not sufficient in most cases. To resolve this problem a new standard was developed over ten years by the International Alliance for Interoperability (IAI).) It is named the IFC format (IFC - Industry Foundation Classes) [8]. The specification is a neutral data format to describe exchange and share information typically used within the building and facility management industry. This norm considers the building elements as independent objects where each object is characterized by a 3D representation and defined by a semantic normalized label. Consequently, the architects and the experts are not the only ones who are able to recognize the elements, but everyone will be able to do it, including the system itself. For instance, an IFC "Signal" is not just a simple collection of lines and geometric primitives recognized as a signal; it is an "intelligent" object signal which has attributes linked to a geometrical definition and function. IFC files are made of objects and connections between these objects. Object attributes describe the "business semantic" of the object. Connections between objects are represented by "relation elements" [1].

As a matter of fact, the WiDOP project (knowledge-based detection of objects in point clouds) aims at making a step forward. The goal is to develop efficient and intelligent methods for an automated processing of terrestrial laser scanner data, Figure **1**. The principle of the WiDop project is a knowledge-based detection of objects in point clouds for AEC (Architecture, Engineering and Construction) engineering applications using IFC format. In contrast to existing approaches, the project consists in using prior knowledge about the context and the objects. This knowledge is extracted from databases, CAD plans, Geographic Information Systems (GIS), technical reports or domain experts. Therefore, this knowledge is the basis for a selective knowledge-oriented detection and recognition of objects in point clouds.

The project WiDOP is Funded by the German government. However, the partners are the Fraport company (Frankfurt Airport manager), the German railway company (Deutsche Bahn), and the Metronom company which is specialized in 3D point cloud processing. Where the Deutsche Bahn main concerns are the management of the railway furniture. Actually, the environment of the railway is constantly changing. Where the cost of keeping these plans up to date is increasing. The present-time solution adopted by the Deutsche Bahn (DB) consists on fixing a 3D terrestrial laser scanner on the train and to survey the surrounding landscape (Railway, signals and green trees on the borders). Metronom automation is a DB subcontractor specialized in 3D data processing. This partner takes the survey point clouds as input and detects the different existent elements manually helped with some 3D process like spike detection. The main objective of Deutsch Bahn project consists in detecting automatically the objects in the 3D point clouds to feed the position and the semantic definition of objects into a GIS system.

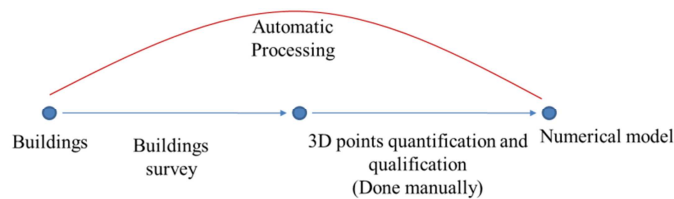

Figure 1. Automatic processing compared to the manual one

The present project aims at building a bridge between the semantic modeling and the numerical processing, to define strategies based on domain knowledge and 3D processing knowledge. The knowledge will be structured in ontologies containing a variety of elements like already existing information about objects of that scene. Like data sources (digital maps, geographical information systems, etc.), information about the objects' characteristics, the hierarchy of the sub-elements, the geometrical topology, the characteristics of processing algorithms, etc. In addition, all relevant information about the objects, geometries, inter and intra-relation and the 3D processing algorithms have been modeled inside the knowledge base, including characteristics such as positions, geometrics information, images textures, behavior and parameter of suitable algorithms, for example. The suggested system is materialized via WiDOP project [6]. Furthermore, the created WiDOP platform can generate an indexed scene from unorganized 3D point clouds visualized within the virtual reality modeling language. [7].

II. BACKGROUND CONCEPT AND METHODOLOGY

The problematic of 3D object detection and scene reconstruction, including semantic knowledge was recently treated within different domains. Basically, the photogrammetry one [9], the construction one, the robotics [10] and recently the knowledge engineering one [11]. Modeling a survey, in which low-level point cloud or surface representation is transformed into a semantically rich model is done through three main tasks. T first is the data collection, in which dense point measurements of the facility are collected using laser scans taken from key locations throughout the facility; Then data processing, in which the sets of point clouds from the collected scanners are processed. Finally, modeling the survey in which the low-level point cloud is transformed into a semantically rich model. This is done via modeling geometric knowledge, qualifying topological relations and finally assigning an object category to each geometry [12]. Concerning the geometry modeling, we remind here that the goal is to create simplified representations of facility components by fitting geometric primitives to the point cloud data. The modeled components are labeled with an object category. Establishing relationships between components is important in a facility model and must also be established. In fact, relationships between objects in a facility model are useful in many scenarios. In addition, spatial relationships between objects provide contextual information to assist in object recognition [13]. Within the literature, three main strategies are described to rich such a model where the first one is based on human interaction with provided software's for point clouds classifications and annotations [14]. While the second strategy relies more on the automatic data processing without any human interaction by using different segmentation techniques for feature extraction [10]. Finally, new techniques presenting an improvement compared with the cited ones by integrating semantic networks to guide the reconstruction process are presented in [15].

### A. Manual survey model creation

In current practice, the creation of a facility model is largely a manual process, performed by service providers who are contracted to scan and model a facility. In reality, a project may require several months to be achieved, depending on the complexity of the facility and the modeling requirements. Reverse engineering tools excel at geometric modeling of surfaces, but with the lack of volumetric representations, while such design systems cannot handle the massive data sets from laser scanners. As a result, modelers often shuttle intermediate results back and forth between different software packages during the modeling process, giving rise to the possibility of information loss due to limitations of data exchange standards or errors in the implementation of the standards within the software tools [16]. Prior knowledge about component geometry, such as the diameter of a column, can be used to constrain the modeling process, or the characteristics of known components may be kept in a standard component library. Finally, the class of the detected geometry is determined by the modeler once the object is created. In some cases, relationships between components are established either manually or in a semi-automated manner.

### B. Semi-Automatic and Automatic methods

The manual process for constructing a survey model is time consuming, labor-intensive, tedious, subjective, and requires skilled workers. Even if modeling of individual geometric primitives can be fairly quick, modeling a facility may require thousands of primitives. The combined modeling time can be several months for an average-sized facility. Since the same types of primitives must be modeled throughout a facility, the steps are highly repetitive and tedious [17]. The above-mentioned observations and others illustrate the need for semi-automated and automated techniques for facility model creation. Ideally, a system could be developed that would take a point cloud of a facility as input and produce a fully annotated as-built model of the facility as output. The first step within the automatic process is the geometric modeling. It presents the process of constructing simplified representations of the 3D shape for survey components from point cloud data. In general, the shape representation is supported by CSG [18] or B-Rep [19] representation. The representation of geometric shapes has been studied extensively [20]. Once geometric elements are detected and stored via a specific presentation, the final task within a facility modeling is the object recognition. It presents the process of labeling a set of data points or geometric primitives extracted from the data with a named object or object class. Whereas the modeling task would find a set of points to be a vertical plane, the recognition task would label that plane as being a wall, for instance. Often, the knowledge describing the shapes to be recognized is encoded in a set of descriptors that implicitly capture object shape. Research on recognition of facilities specific components is still in its early stages. Methods in this category typically perform an initial shape-based segmentation of the scene, into planar regions, for example, and then use features derived from the segments to recognize objects. This approach is exemplified by Rusu et al. who use heuristics to detect walls, floors, ceilings, and cabinets in a kitchen environment [10]. A similar approach was proposed by Pu and Vosselman to model facility façades [21].

To reduce the search space of object recognition algorithms, the use of knowledge related to a specific facility can be a fundamental solution. For instance, Yue et al. overlay a design model of a facility with the as-built point cloud to guide the process of identifying which points clouds data belong to specific objects and to detect differences between the as-built and as-designed conditions [22]. In such cases, object recognition problem is simplified to be a matching problem between the scene model entities and the data points. Another similar approach is presented in [23]. Other promising approaches have only been tested on limited and very simple examples, and it is equally difficult to predict how they would fare when faced with more complex and realistic data sets. For example, the semantic network methods for recognizing components using context work well for simple examples of hallways and barren, rectangular rooms [13], but how would they handle spaces with complex geometries and clutter.

### C. Discussion

The presented methods for survey modeling and object recognition rely on knowledge about the domain. Concepts like "Signals are vertical" and "Signals intersect with the ground" are encoded explicitly through a set of rules. Such rule based approaches tend to be brittle and break down when they are tested in new and slightly different environments. Additionally, regarding the literature, people models the context by specifying the concepts and the relationships of objects to describe the world. However, no one mentions the knowledge about the 3D processing algorithms and the associated results such as the geometry and the topology.

Based on these observations, flexible representations of facility objects and more sophisticated guidance based algorithms for object detection by modeling *algorithmic*, *geometric* and *topological knowledge* within an ontology structure the way of a significant improvement. Actually, it will allow the process to create a dynamic sequence of 3D processing algorithms for object detections and to guarantee an automatic detection and recognition of objects in 3D point clouds, materialized via the semantic annotation process.

III. OVERVIEW OF THE WiDOP GENERAL MODEL

In general, mathematical algorithms contain different data processing steps, which are combined with internal decisions based on numerical results. This makes the processing inflexible and error prone, especially when the data does not behave as the model behind the algorithm expects. One of the purposes behind this contribution is to put these implicit decisions outside, make a semantic layer out of it and combine it with the object model. This approach is more flexible and can be easily extended, since knowledge and data processing are separated.

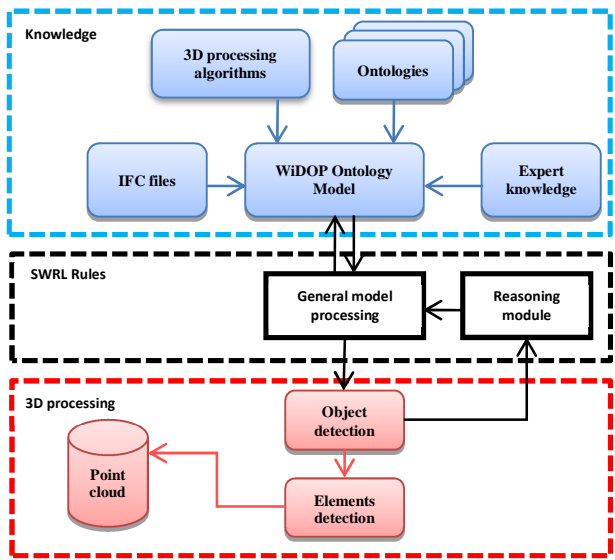

Figure 2. WiDOP: Overview system

Figure 2 presents the general architecture for the WiDOP project. It is composed of three parts: the knowledge model, the 3D processing algorithms execution, and the interaction management and control part labeled WiDOP processing materialized within rules and extensions, ensuring the interaction between the above sited parts. In contrast with existing approaches, we aim at the utilization of previous knowledge on objects. This knowledge can be contained in databases, construction plans, as-built plans or Geographic Information Systems (GIS).

A. The knowledge model

The term "Semantic Web" has been defined numerous time. Though there is no formal definition of Semantic Web, some of its most used definitions are "The Semantic Web is not a separate Web but an extension of the current one, in which information is given well-defined meaning, better enabling computers and people to work in cooperation" [28]. It is a source to retrieve information from the Web (using the Web spiders from RDF files) and access the data through Semantic Web Agents or Semantic Web Services. Simply, Semantic Web is data about data or metadata. "A Semantic Web is a Web where the focus is placed on the meaning of words, rather than on the words themselves, where information becomes knowledge after semantic analysis is performed. For this reason, a Semantic Web is a network of knowledge, compared with what we have today, that can be defined as a network of information [29]. The Semantic Web provides a common framework that allows data to be shared and reused across application, enterprise and community boundaries [30]. In fact, description logics provide a formalization for knowledge representation of real-world situations. This provides the logical replies to the queries of real-world situations. The results are highly sophisticated reasoning engines, which utilize the expressiveness capabilities of DLs to manipulate the knowledge. A Knowledge Representation system is a formal representation of a knowledge described through different technologies. When it is described through DLs, they set up a Knowledge Base (KB), the contents of which could be reasoned or infer to manipulate them. A knowledge base could be considered as a complete package of knowledge content. It is, however, only a subset of a Knowledge Representation system that contains additional components.

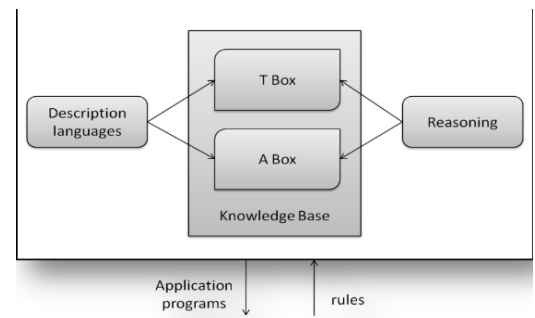

Figure 3. The Architecture of a knowledge representation system

As seen in Figure 3, the author [31] sketches the architecture of any Knowledge Representation system based on DLs. It could be seen the central theme of such a system is a Knowledge Base (KB). It is composed by two components: the TBox and the ABox. TBox statements are the terms or the terminologies that are used within the system domain. In general, they are statements describing the domain through the controlled vocabularies. For example, in terms of a Deutsche Bahn domain the TBox statements are the set of concepts as Signal, Fourniture, ProcessingAlgorithm, etc. or the set of roles as hasCharacteristics, isDeseignedFor, hasGeometry etc. ABox in contains assertions to the TBox statements. For example, Wall1 is an ABox presents the TBox Wall.

Our approach is intended to use semantics based on OWL technology [44] for knowledge modeling and processing. Knowledge has to be structured and formalized based on IFC schema, XML files and particularly on Deutsche Bahn and 3D processing domain experts, etc., using classes, instances, relations and rules. An object in the ontology can be modeled as presented; a room has elements composed of walls, a ceiling and a floor. The sited elements are basic objects. They are defined by their geometry (plane, boundary, etc.), features (roughness, appearance, etc.), and also the qualified relations between them (adjacent, perpendicular, etc.). The object "room" gets its geometry from its elements, where further characteristics may be added, such as functions in order to estimate the existent sub elements. For instance, a "classroom" will contain "tables", "chairs", "a blackboard", etc. The research of the object "room" will be based on an algorithmic strategy which will look for the different objects contained in the point cloud. This means, using different detection algorithms for each element, based on the above mentioned characteristics, will allow us to classify most of the point region in the different element categories. It corresponds to the spatial structure of any facility, and it is an instance of semantic knowledge defined in the ontology. This instance defines the rough geometry and the semantics of the building elements without any real measurement. This model contains also knowledge extracted from the technical literature of the domain and knowledge from experts of the domain also. In addition, the ontology is, as well enriched with knowledge about 3D processing algorithms and populated with the results of experiences undertaken on 3D point clouds, which define the empirical knowledge extracted from point clouds regarding a specific domain of application.

*B. The 3D processing algorithms*

Numerical processing includes a number of algorithms or their combination to process the spatial data. Strategies include geometric element detection (straight line, plane, surface, etc.), projection-based, region estimation, histogram matrices, etc. All of these strategies are either under the guidance of knowledge, or use the previous knowledge to estimate the object intelligently and optimally. Alongside with 3D point clouds, various types of input data sets can be used such as images, range images, point clouds with intensity or color values, point clouds with individual images oriented to them or even stereo images without a point cloud. All sources are exploited for application to particular strategies. Knowledge not only describes the information of the objects, but also gives a framework for the control of the selected strategies. The success rate of detection algorithms using RANSAC [24], Iterative Closest Point [25] and Least Squares Fitting [26] should significantly increase by making use of the knowledge background. However, we are planning not only to process point data sets, but also surface and volume representation like mesh, voxels and bounding Boxes. These methods and others will be selected in a flexible way, depending on the semantic context.

*C. The WiDOP processing*

In order to manage the interaction between the knowledge part and the 3D processing part, a new layer labeled WiDOP processing materialized within rules is created. This layer ensures the control and the management of the knowledge transaction and the decision taken based on SWRL languages, and its extensions through several steps explained in the next section. The semantic within the ontologies expressed through OWL can be used inside the ontologies, and the knowledge bases themselves for inference purposes. However, in order to express the rules, the Semantic Web Rule Language (SWRL) is emerged [45]. The SWRL has the form antecedent → consequent, where both antecedent and consequent are conjunctions of atoms written $a_1 \wedge ... \wedge a_n$. Atoms in swrl rules can be of the form $C(x)$, $P(x,y)$, $Q(x,z)$, sameAs(x,y), differentFrom(x,y), or builtIn (pred, $z_1$, …, $z_n$), where C is an OWL description, P is an OWL individual-valued property, Q is an OWL data-valued property, pred is a datatype predicate, x and y are either individual-valued variables or OWL individuals, and z, $z_1$, … $z_n$ are either data-valued variables or OWL data literals. An OWL data literal is either a typed literal or a plain literal. Variables are indicated by using the standard convention of prefixing them with a question mark (e.g., ?x). URI references (URIrefs) are used to identify ontology elements such as classes, individual-valued properties and data-valued properties. For instance, the following rule asserts that one's parents' brothers are one's uncles where parent, brother and uncle are all individual-valued properties.

parent(?x, ?p) ∧ brother(?p, ?u) → uncle(?x, ?u)     (1)

The set of built-ins for SWRL are motivated by a modular approach allowing further extensions in future releases within a taxonomy. SWRL's built-ins approach is also based on the reuse of existing built-ins in XQuery or XPath, which are themselves based on XML Schema by using Datatypes. The system of built-ins should as well help in the interoperation of SWRL with other Web formalisms, by providing an extensible, modular built-ins infrastructure for Semantic Web Languages, Web Services, and Web applications. Many built-ins are defined. These built-ins are keys for any external integration. This project takes advantages of this extensional mechanism to integrate new Built-ins for 3D processing and topological processing.

*D. Interaction process*

To focus on our method for the combination of the Semantic Web technologies and the 3D processing algorithms, Figure 4 illustrates an UML sequence diagram that represents the general design of the proposed solution. Hence, the purpose is to create a more flexible, easily extended approach where algorithms will be executed reasonably and adaptively on particular situations following an interaction process.

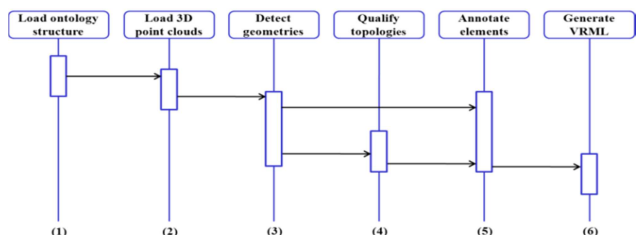

Figure 4. The sequence diagram of interactions between the laser scanner, the 3D processing, the knowledge processing and the knowledge base.

The processing steps can be detailed where three main steps aim at detecting and identifying objects.

(3) From 3D point clouds to geometric elements.
(4) From geometry to topological relations.
(5) From geometric and/or topological relations to semantic elements annotated.

As intermediate steps, the different geometries within specific 3D point clouds are detected and stored in the ontology structure. Once done, the existent topological relations between the detected geometries are qualified and then populated within the prior knowledge. Finally, detected geometries are annotated semantically, based on existing knowledge's related to the geometric characteristics and topological relations. The input ontology contains knowledge about the Deutsche Bahn railway objects and knowledge about 3D processing algorithms.

IV. DESCRIPTION OF THE WiDOP KNOWLEDGE BASE

This section discusses the different aspects related to the Deutsche Bahn scene ontology structure installed behind the WiDOP Deutsche Bahn prototype [11]. The domain ontology presents the core of WiDOP project and provides a knowledge base to the created application. The global schema of the modeled ontology structure offers a suitable framework to characterize the different Deutsche Bahn elements from the 3D processing point of view. The created ontology is used basically for two purposes:

- To guide the processing algorithm sequence creation based on the target object characteristics.
- To facilitate the semantic annotation of the different detected objects inside the target scene.

The current ontology, following to above considerations and with respect to technological possibilities, will be modeled in various levels. In principle, we have to distinguish between object-related knowledge and algorithmic related knowledge. In addition, the same distinction has to be done on the *layer of the object knowledge* and the *layer of the algorithmic knowledge* containing the respective semantic information. In fact, the ontology is managed through different components of description logics where we find five main classes within other data and objects properties able to characterize the scene in question.

- Algorithm
- Geometry
- DomainConcept
- Characteristics
- Scene

The *DomainConcept* class can be considered as the main class in the ontology as it is the class where the target objects are modeled. However, the importance of other classes cannot be ignored. They are used to either describe the object geometry, through the *Geometry* class by defining its geometric component or the bounding rectangle of the object that indicate its coordinates, or to either describe its characteristics through the *Characteristics* class. Additionally, the suitable algorithms are automatically selected based on its compatibility within the object geometry and characteristics. Add to that, other classes are equally significant but play their roles in the backend. The connection between the basic mentioned classes is carried out through object and data properties. There exist object properties for each mentioned activities. Besides, the object properties are also used to relate an object to other objects via topological relations. In general, there are five general object properties in the ontology which have their specialized properties for the specialized activities, Figure 5. They are:

- hasTopologicRelation
- IsDeseignedFor
- hasGeometry
- hasCharacteristics

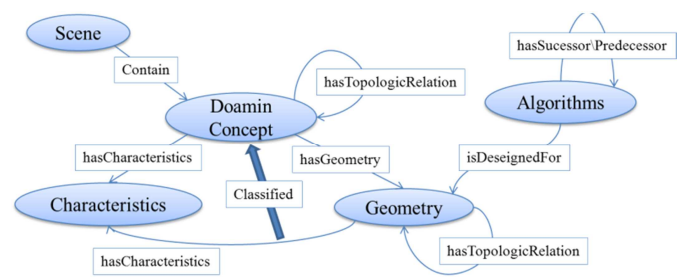

Figure 5. Ontology general schema overview

The next sections focus on the layers, the object and the algorithmic knowledge definition.

*A. Layers of object knowledge*

The object knowledge layer will be classified in three categories: geometric, topological and semantic knowledge representing a certain scenario [35]. Therefore we distinguish between:
- Deutsche Bahn Scene knowledge
- Geometric knowledge
- Topological knowledge

*1) Layer of the Deutsche Bahn Scene knowledge*

The layer of object knowledge contains all relevant information about objects and elements which might be found within a Deutsch Bahn scene. This might comprise a list such as: {Signals, Mast, Schalanlage, etc.}. They are used to fix either the main scene within its point clouds file and its size through attributes related to the scene class, or even to characterize detected element with different semantic and geometric characteristics.

The created knowledge base related to the Deutsche Bahn scene has been inspired next to our discussion with the domain expert and next to our study based on the official Web site for the German rail way specification [46]. An overview of the targeted elements, the most useful and discriminant characteristics to detect it and their inter-relationship is presented.

Table 1. EXAMPLE OF THE DB SCENE OBJECTS

| Class | Sub Class | Subsub Class | Height | Correspondent image |
|---|---|---|---|---|
| Signals | Basic Signals | Main Signal | Between 4 and 6 m | 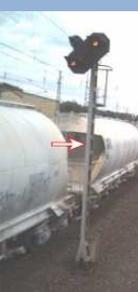 |
| | | Distant Signal | Between 4 and 6 m | |
| | Secondary signal | Vorsignalbake | between 1,5 and 2.5 m | |
| | | Breakpoint_table | between 1 and 2 m | |
| | | Chess_board | between 1 and 1,5 m | |
| Mast | BigMast | | More than 6m | 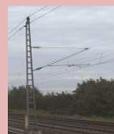 |
| | NormalMast | | Between 5 and 6 | |
| Schaltanlage | Schalthause | | Less than 1m | 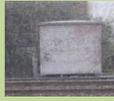 |
| | ScheltSchrank | | Less than 0,5m | |

Table **1** shows a possible collection of scene elements in case of a Deutsche Bahn scene. They may be additionally structured in a hierarchical order as might be seen convenient for a scene, while Figure 6 shows the suggested structure to model them within the OWL language.

Basically, a railway signal is one of the most important elements within the Deutsche Bahn scene, where we find main signals and secondary ones. The main signals are classified onto *the primary signal* and the *distant ones*. In fact, the primary signal is a railway signal. It indicates whether the subsequent section of track may be driven on. A primary signal is usually announced through a distant signal. The last one indicates which image signal to be expected, that will be associated to the main signal in a distance of 1 km. Big variety of secondary signals exists like the Vorsignalbake, the Haltepunkt and others. From the other side, the other discriminant elements within the same scene are the Masts presenting electricity born for the energy alimentation. Usually, masts are distant from 50 m to each other. Finally, the Schaltanlage elements present small electric born connected to the ground.

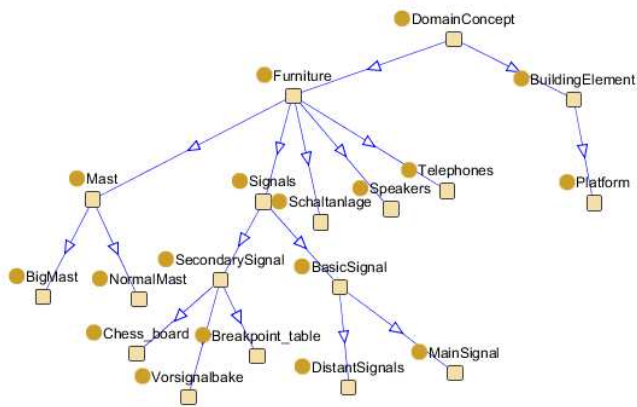

Figure 6. Example of the DB scene objects modeling

Additionally, the above cited concepts are extended by relations to other classes or data. As an example, the data property "*has_Bounding_Box*" aims to store the placement of the detected object in a bounding box defined by its eight 3D points (each 3D point is defined by three values x, y and z).

To specify its semantic characteristics, new classes are created, aiming to characterize a semantic object by a set of characteristics like color, size, visibility, texture, orientation and its position in the point cloud after detection. To do so, new object properties like "*has_Color*", "*has_Size*", "*has_Orientation*", "*has_Visibility*" and "*has_Texture*" are created linking the *Semantic_Object* class to the "*color*", "*size*", "*Orientation*", "*Visibility*" and "*Texture*" classes respectively.

*2) Layer of the geometric knowledge*

Geometrical knowledge formulates geometrical characteristics to the physical properties of scene elements. In the simplest case, this information might be limited to few coordinates expressing a bounding box containing the object. However, for elements being accessible to functional descriptions, additional knowledge will be mentioned. A signal, for example, has vertical lines, which needs to be described by a line equation, its values and completed by width and height. In fact, we think that such knowledge can present a discriminant feature able to improve the automatic annotation process. For this reason, we opt to study the different geometric features related to the cited semantic elements, then, use only the discriminant one as basic features for a given object. The following table gathers the object characteristics together regarding the properties of a bounding box,

Table **2**, Figure 6. This table is extended with algorithm characteristics, but it is not presented here.

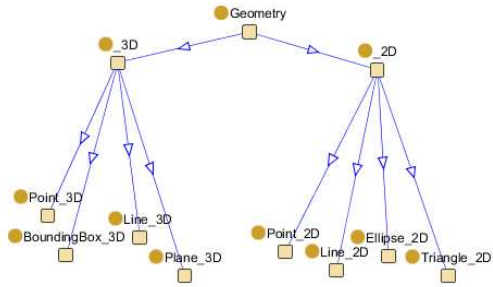

Figure 7. The geometry class hierarchy

Table 2. Geometric characteristics overview

| Class | SubClass | Subsub Class | Restriction on Line number | Restriction on Planes number |
|---|---|---|---|---|
| Signals | Basic Signals | Main Signal | 1 or 2 Vertical line | 0 |
| | | Distant Signal | 1 or 2 Vertical line | 0 |
| | Secondary signal | Vorsignalbake | 1 Vertical line | 1 Vertical plane |
| | | Breakpoint_table | 2 Vertical lines | 1 Vertical Plan |
| | | Chess_board | 1 Vertical line | 1 Vertical plane |
| Mast | BigMast | More than 6m | 2 or 4 vertical lines | 0 |
| | NormalMast | Between 5 and 6 | 2 or 4 vertical lines | 0 |
| Schaltanlage | Schalthause | Less than 1m | | 1 Vertical plane 1 Horizontal plane |
| | SchaltSchrank | Less than 0,5m | | 1 vertical plane |

*3)* Layer of the topological knowledge

While exploring the railway domain, lots of standard topological rules are imposed; such rules are used to help the driver and to ensure the passengers' security. From our point of view, these are helpful also to verify and to guide the annotation process. In fact, topological knowledge represents adjacency relationships between scene elements. For instance, and in case of the Deutsche Bahn scene, the distance between the distant signal and the main one corresponds to the stopping distance that the trains require. The stopping distance shall be set on specific route and is in the main lines often 1000 m or in a rare case, 700 m. Add to that, three to five Vorsignalbake are distant from 75m while then the last one is distant from 100m to the distant signal, Figure 8.

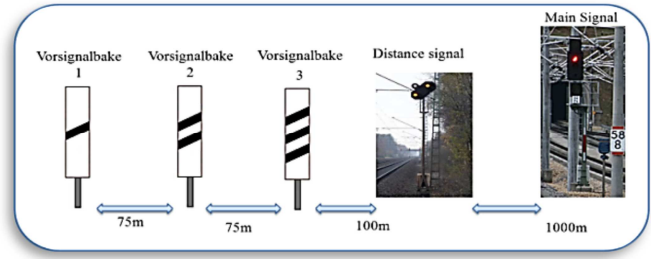

Figure 8: Topologic rules

The purpose of such object properties is to spatially connect *Things* presented in the scene. At semantic view, topological properties describe adjacency relations between classes. For example, the property isParallelTo allows characterizing two geometric concepts by the feature of parallelism. Similarly relations like isPerpendicularTo and isConnectedTo will help to characterize and exploit certain spatial relations and make them accessible to reasoning steps.

*B. Layer of processing knowledge*

The 3D processing algorithmic layer contains all relevant aspects related to the 3D processing algorithms. It´s integration into the semantic framework is done by special Built-Ins called "Processing Built-Ins". They manage the interaction between above mentioned layers. In addition, it contains algorithm definitions, properties, and geometries related to each defined algorithms. An importance achievement is the detection and the identification of objects, which has a linear structure such as signal, indicator column, and electric pole, etc., through utilizing their geometric properties. Since the information in point cloud data sometimes is unclear and insufficient, the various methods to RANSAC [24] are combined and upgraded. This combination is able to robustly detect the best fitting lines in 3D point clouds for example.

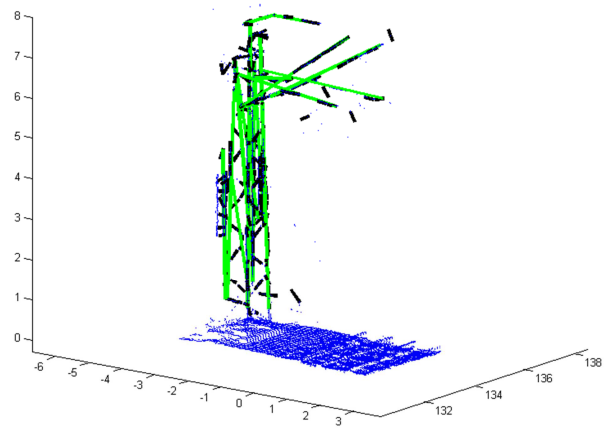

Figure 9. Mast detection

Figure 9 contains the Mast object constructed by linear elements, ambiguously represented in point cloud as blue points. Green lines are results of possible fitting lines and clearly show the shape of the object that is defined in the ontology. The object generated from this part is a bounding box that includes all inside geometries of the object and a concept label.

Next to the 3D expert recommendation, knowledge within the Table 3 is created, where a set of 3D processing algorithms within the target detected geometry are structured; the input and output are created.

Table 3. 3D Algorithms and experts observations

| Algorithm name | has Input | hasOutput | isDesignedfor | hasSuccessor |
|---|---|---|---|---|
| Vertical Objects Detection | PointCloud | Point_2D | Vertical gemetry | None |
| Segmentationin2D | Point_2D PointCloud | SubPointCloud | Vertical gemetry | VerticalObjectsDetection |
| BoundingBox | SubPointCloud | Point_3D | Vertical gemetry | Segmentationin2D |
| Approximate Height | SubPointCloud | number | Geometry height | Segmentationin2D |
| RANSAC Line Detection | SubPointCloud | Line_3D | 3D Lines | Segmentationin2D) |
| FrontFaceDetection | SubPointCloud | Boolean | Geometry with front face | Segmentationin2D |
| CheckPerpendicular | Line_3D | Boolean angle | Geometry containing Perpendicular elements | LinesDetectionin3DbyRANSAC |
| CheckParallel | Line_3D | Boolean angle | Geometry containing Parallel elements | LinesDetectionin3DbyRANSAC |

The subclasses of the *Algorithm* class, Figure 10, are representing all the algorithms developed in the 3D processing layer. They are related to several properties which they are able to detect. These properties (Geometric and semantic) are shared with the *DomainConcept* and the *Geometry* classes. By this way, a sequence of algorithms can detect all the characteristics of an element.

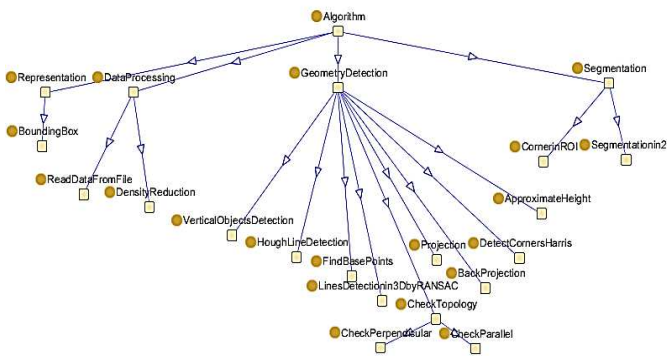

Figure 10. Hierarchical structure of the Algorithm class

The next section introduces an overview of the approach undertaken in the WiDOP project to detect and annotate semantically the different Deutsch Bahn objects.

V. INTELLIGENT PROCESS

The basic strength of formal ontology is their ability to reason in a logical way based on Descriptive Logic language DL [36]. The last one presents a form of logic to reason on objects. Lots of reasoners exist nowadays like Pellet [37], and KAON [38]. Actually, despite the richness of OWL's set of relational properties, the axioms does not cover the full range of expressive possibilities for object relationships that we might find, since it is useful to declare relationship in term of conditions or even rules. These rules are used through different rules languages to enhance the knowledge possess in an ontology.

Within the WiDOP project, the domain ontologies are used to define the concepts, and the necessary and sufficient conditions that describe the concepts. These conditions are of value, because they are used to populate new concepts. For instance, the concept "Vertical_BoudinBox" can be specialized into "Signal" if it contains a "VerticalLines". Consequently, the concept "Signal" will be populated with all "Vertical_BoudinBox" if they are linked to a "VerticalLines" with certain parameters. In addition, the rules are used to compute more complex results such as the topological relationships between objects. For instance, the relations between two objects are used to get new efficient knowledge about the object. The ontology is than enriched with this new relationship. The topological relation built-ins are not defined in the SWRL language. Consequently, the language was extended.

To support the defined use cases, two basic further layers to the semantic one are added to ontology in order to ensure the geometry detection and annotation process tasks. These operations are the 3D processing and topological relations qualification respectively.

*A. Integration of 3D processing operations*

The 3D processing layer contains all relevant aspects related to the 3D processing algorithms. Its integration into the WiDOP semantic framework is done by special Built-Ins. They manage the interaction between processing layers and the semantic one. In addition, it contains the different algorithm definitions, properties, and the related geometries to the each defined algorithms. An importance achievement is the detection and the identification of objects with specific characteristics such as a signal, indicator columns, and electric pole, etc. through utilizing their geometric properties. Since the information in point cloud data sometimes is unclear and insufficient.

The Semantic Web Rule Language within extended built-ins is used to execute a real 3D processing algorithm, and to populate the provided knowledge within the ontology (e.g. Table 4). The "3D_swrlb_Processing: VerticalElementDetection" built-ins for example, was created, it aims at the detection of geometry with vertical orientation. The prototype of the designed Built-in is:

*3D_swrlb_Processing:VerticalElementDetection(?Vert, ?Dir)*

where the first parameter presents the target object class, and the last one presents the point clouds' directory defined within the created scene in the ontology structure. At this point, the detection process has as a result a set of bounding boxes, representing a rough position and orientation of the detected object. Table 4 shows the mapping between the 3D processing built-ins, which is computer and translated to predicate, and the corresponding class.

TABLE 4. 3D PROCESSING BUILT-INS MAPPING

| 3D Processing Built-Ins | Correspondent Simple class |
|---|---|
| 3D_swrlb_Processing: VerticalElementDetection (?Vert,?Dir) | Vertical_BoundingBox(?x) |
| 3D_swrlb_Processing: HorizentalElementDetection (?Vert,?Dir) | Horizental_BoundingBox(?y) |

### B. Integration of Topologic operations

The layer of the topological knowledge represents topological relationships between scene elements since the object properties are also used to link an object to others by a topological relation. For instance, a topological relation between a distant signal and a main one can be defined, as both have to be distant from one kilometer. The qualification of topological relations into the semantic framework is done by new topological Built-Ins.

This step aims at verifying certain topology properties between detected geometries. Thus, 3D_Topologic built-ins have been added in order to extend the SWRL language. Topological rules are used to define constrains between different elements. After parsing the topological built-ins and its execution, the result is used to enrich the ontology with relationships between individuals that verify the rules. Similarly to the 3D processing built-ins, our engine translates the rules with topological built-ins to standard rules, Table 5.

TABLE 5. EXAMPLE OF TOPOLOGICAL BUILT-INS

| Processing Built-Ins | Correspondent object property |
|---|---|
| 3D_swrlb_Topology:Upper(?x, ?y) | Upper(?x,?y) |
| 3D_swrlb_Topology:Intersect(?x, ?y) | Intersect (?x,?y) |

### C. Guiding 3D processing algorithms

Actually, the created knowledge base aims to satisfy to basic purposes which are:

- Guiding the processing algorithm sequence creation based on the target object characteristics.
- Facilitate the semantic annotation of the different detected objects inside the target scene.

Let's remember that the one of the main ideas behind this project is to direct, adapt and select the most suitable algorithms based on the object's characteristics. In fact, one algorithm could not detect and recognize different existent objects in the 3D point clouds, since they are distinguished by different shapes, size and capture condition. The role of knowledge is to provide not only the object's characteristics (shape, size, color...) but also object's status (visibility, correlation) to algorithmic part, in order to adjust its parameters to adapt with a current situation. Based on these observations, we issue a link from algorithms to objects based on the similar characteristics as Figure 11 shows.

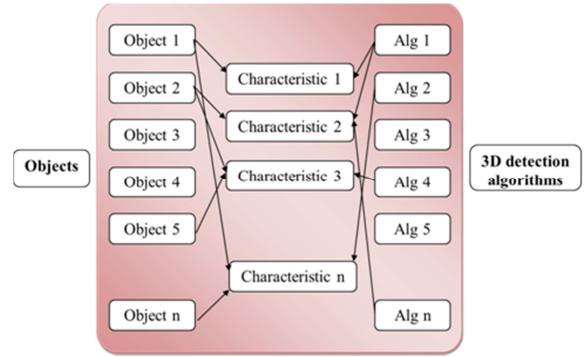

Figure 11. Algorithms selection based on object's characteristics

In fact, knowledge controls one or more algorithms for detecting object. To do this, we try to find a match between the object's characteristics and characteristics that a certain algorithm can be used for. For example, object O has characteristics: $C_1$, $C_2$, $C_3$; and algorithm $A_i$ can detect characteristic $C_1$, $C_3$, $C_4$, while algorithm $A_j$ can detect characteristic $C_2$, $C_5$. Then, decision algorithm will select $A_i$ and $A_j$ since these algorithms have capability detecting the characteristics of object O. The set of characteristics are determined by the object's properties such as geometrical features and appearance. Once done, selected algorithms will be executed and target characteristics will be detected.

The whole process takes as input the 3D point clouds scenes, an ontology structure presenting a knowledge base to manipulate objects, geometries, topologies and relations (Object and data property) and produces as an output, an annotated scene within the same ontology structure. As intermediate steps, the different geometries within a specific 3D point cloud scene are detected and stored in the ontology. Once knowledge about geometries and the topologies are experienced, SWRL rules aim at qualifying and annotating the different detected geometries. The following simple example shows how a SWRL rule can specify the class of a VerticalBoundingBox which is of type Mast regarding its altitude. The altitude is highly relevant only for this element.

```
3DProcessing_swrlb:VerticalElementDetection(?
Vert,  ?dir)  ^   altitude  (?x,   ?alt)
^swrlb:moreThan (?alt, 6) → Mast (?Vert)
```

In other cases, geometric knowledge is not sufficient for the previous process. The topological relationships between detected geometries are helpful to manage the annotation process. The following example shows how semantic information about existing objects is used conjuncty with topological relationships in order to define the class of another object.

```
Mast (?vert1) ^ VerticalBB (?Vert2) ^
hasDistanceFrom (?vert1,?vert2, 50) →
Mast(?vert2)
```

## VI. WiDOP PROTOTYPE

WiDOP prototype takes in consideration the adjustment of the old methods and, in the meantime, profit from the advantages of the emerging cutting-edge technology. From the principal point of view, our system still retains the storing mechanism within the existent 3D processing algorithms; in addition, suggest a new field of detection and annotation, where we are getting a real-time support from the target scene knowledge. Add to that, we suggest a collaborative Java Platform based on semantic web technology (OWL, RDF, and SWRL) and knowledge engineering in order to handle the information provided from the knowledge base and the 3D packages results.

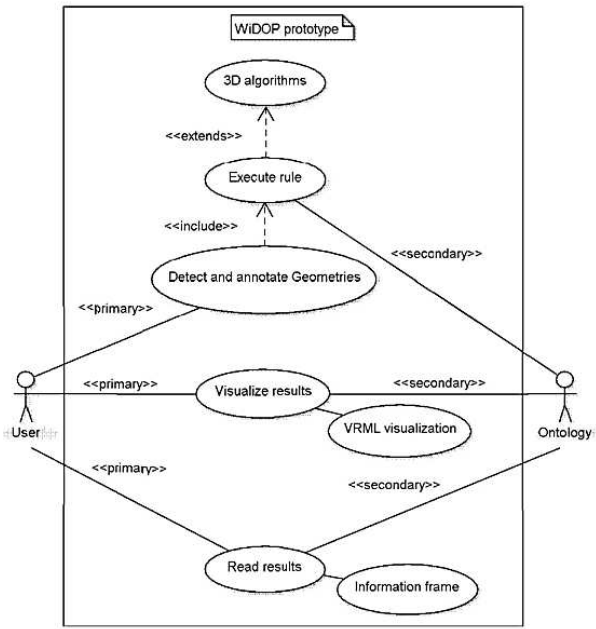

Figure 12. the WiDOP use case diagram

The process enriches and populates the ontology with new individuals and relationships between them. In order to graphically represent these objects within the scene point clouds, a VRML model file [7] is generated and visualized within the prototype where the color of objects in the VRML file represents its semantic definition. The resulting ontology contains enough knowledge to feed a GIS system, and to generate IFC file [37] for CAD software. As seen in Figure 12, the created system is composed of three parts.

- Generation of a set of geometries from a point could file based on the target object characteristics.
- Computation of business rules with geometry, semantic and topological constrains in order to annotate the different detected geometries.
- Generation of a VRML model related to the scene within the detected and annotated elements.

In addition, the created WiDOP platform offers the opportunity to materialize the annotation process by the generation and the visualization based on a VRML structure alimented from the knowledge base. It ensures an interactive visualization of the resulted annotation beginning from the initial state, to a set of intermediate states, coming finally to an ending state, Figure 13 where the set of rules are totally executed.

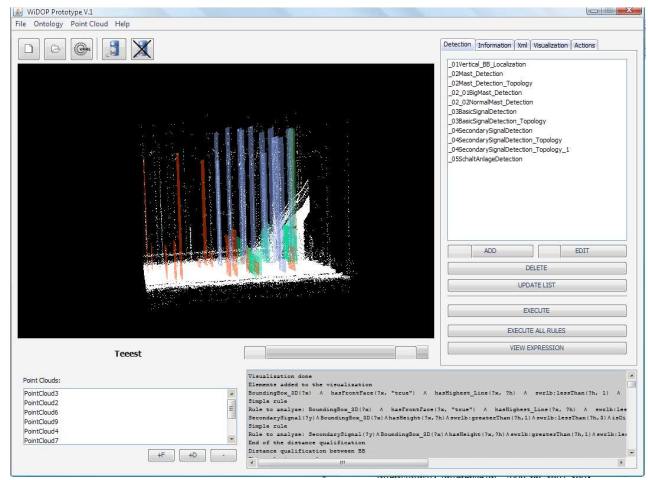

Figure 13. Snapshot of the WiDOP prototype

As a first impression, the system responds to the target requirement since it would take a point cloud of a facility as input and produce a fully annotated as-built model of the facility as output.

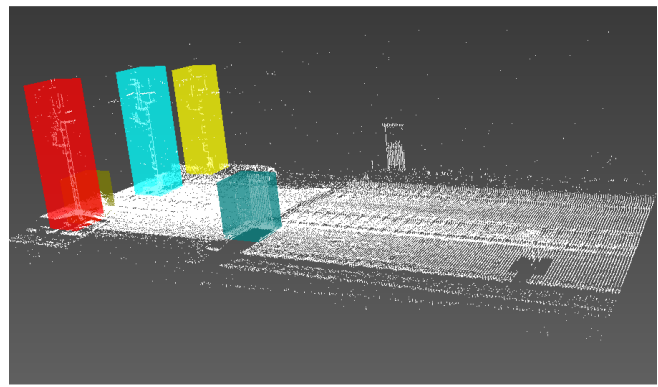

Figure 14. Detected and annotated elements visaliazation within VRML language

## VII. BENCHMARKS

### A. Annotation process summerized

For the demonstration of our prototype, two different sections from the whole scanned point clouds related to Deutsch Bahn scene in the city of *Nürnberg* was extracted. While the last one measure 87 km, we have just taken two small scenes of 500 m each one. Each one of the kept scenes contains a variety of the target objects. The whole scene has been scanned using a terrestrial laser scanner fixed within a train, resulting in a large point cloud representing the surfaces of the scene objects. Within the created prototype, different SWRL rules are processed, e.g. Figure 15. First, geometrical elements will be searched in the area of interest based on dynamic 3D processing algorithm sequence created based on semantic object properties.

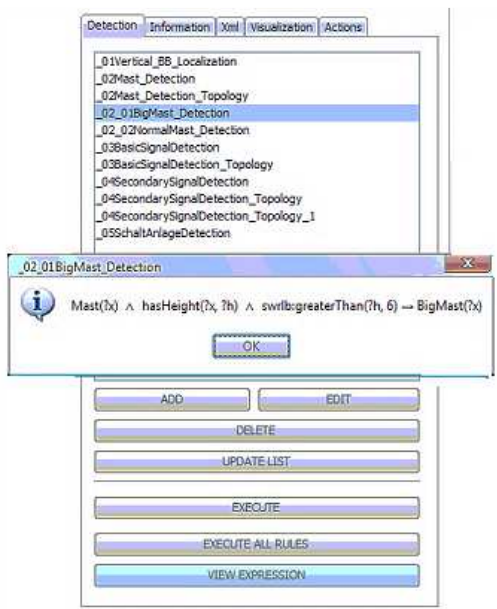

Figure 15. Example of executed rules

Once done, the second step within our approach aims to identify existing topologies between the detected geometries. Thus, useful topologies for geometry annotation are tested. Topological Built-Ins like `isConnected`, `touch`, `Perpendicular`, `isDistantfrom` are created. As a result, relations found between geometric elements are propagated into the ontology, serving as an improved knowledge base for further processing and decision steps.

The last step consists in annotating the different geometries. Vertical elements of certain characteristics can be annotated directly. Subsequently, further annotation may be relayed on aspects expressing facts to orientation or size of elements, which may be sufficient to finalize a decision upon the semantic of an object or, in more sophisticated cases, our prototype allows the combination of semantic information and topological ones that can deduce more robust results by minimizing the false acceptance rate, Figure 16. By this way, and based on a list of SWRL rules, most of the detected geometries are annotate as seen within Table 6, Table 7, Table8.

### B. System evaluation

Our testes had been made on two different data bases with 500 m long extracted from the whole scanned point clouds data. Where the first scene contains just 37 elements, and the second one contain 128 elements. As a first impression, it´s totally reasonable that the number of elements varies from a scene to another, because we are near from the rail way station, more the scene is rich and vice versus. It's also clear from the above-mentioned tables, how our knowledge base could recognize which geometry represents a real element from those which are noise, Table 6.

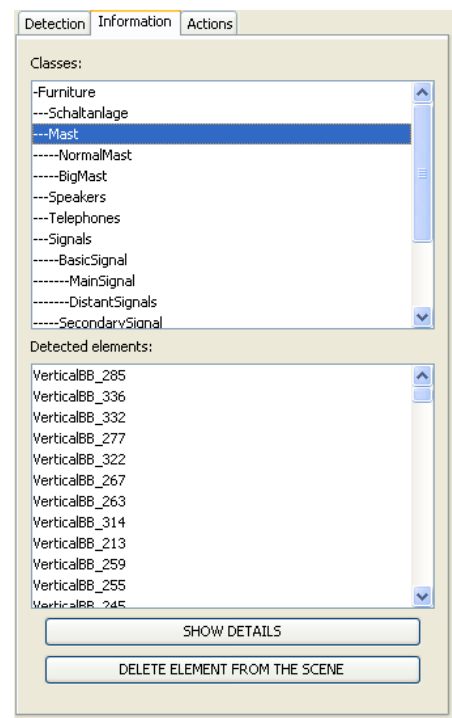

Figure 16. Annotated Bounding Box as Masts

As well, in most cases, our annotation process is able to affect the right label to the detected Bounding box based on knowledge on its component, its internal and external topology. In the first example, Table 7, among 13 elements are classified as Masts, three as a SchaltAnlage and 18 signals. While in the second scene, among 67 elements are classified as Masts, 55 signals and finally 155 Schaltanlage, Table 8.

TABLE 6. DETECTED ELEMENT WITHIN THE SCENE AND ANNOTATED ONES

|  | Scene Size | Detected Bounding Box | Annotated elements | Truth data |
|---|---|---|---|---|
| **Scene1** | 500m | 105 | 34 | 37 |
| **Scene2** | 500m | 344 | 277 | 128 |

TABLE 7. DETECTED AND ANNOTATED ELEMENTS WITHIN THE SCENE1

|  | Masts | signal | Schaltanlage |
|---|---|---|---|
| **Annotated** | 13 | 18 | 3 |
| **Truth data** | 12 | 20 | 5 |

TABLE 8. DETECTED AND ANNOTATED ELEMENTS WITHIN THE SCENE2

|  | Masts | signal | Schaltanlage |
|---|---|---|---|
| **Annotated** | 67 | 55 | 155 |
| **Truth data** | 65 | 50 | 13 |

Some clear limits are detected within the Table 8. Where lots of false Schaltanlage are detected and annotated. Before explaining the reason behind this false detection, let's recall that the Schaltanlage present very small electronic boxes installed on the ground. In the case of scene 2 which is near the rail station, the level of the ground is a higher compared to the other scenes. For this reason, lots of bounding boxes are detected where a high average of them presents small noise on the ground. The reason for the false annotation is the lack of semantic characteristics related to such elements because, until now; there is no real internal or external topology neither internal geometric characteristic that discriminate such an element compared to others.

## VIII. DISCUSSION AND CONCLUSION

We have presented an automatic system for survey information model creation based on semantic knowledge modeling. Our solution aims to perform the detection of objects from a technical survey within the laser scanner technology by using available knowledge about a specific domain (DB). This prior knowledge is modeled within an ontology structure. SWRL rules are used to control the 3D processing execution, the topological qualification and finally to annotate the detected elements in order to enrich the ontology and to drive the detection of new objects.

The designed prototype takes 3D point clouds of a facility, and produce fully annotated scene within a VRML model file. The suggested solution for this challenging problem has proven its efficiency through real tests within the Deutsche Bahn scene. The creation of processing and topological Built-Ins has presented a robust solution to resolve our problematic and to prove the ability of the semantic web language to intervene in any domain and create the difference.

Future work will include the integration of new knowledge's that can intervene within the annotation process like the number of detected lines within each bounding box and the update of the general platform architecture, by ensure more communication between the scene knowledge within the 3D processing one. It will also include a more robust identification and annotation process of objects based on individual object characteristics. Finally, further knowledge related to the algorithm parameterization that can intervene within the detection and annotation process will be studied to make the process more flexible and intelligent.

## IX. ACKNOWLEDGMENT

This paper presents work performed in the framework of the research project funded by the German ministry of research and education under contract No. 1758X09. The authors cordially thank for this funding. Special thinks also for Andreas Marbs, Ashish Karmacharya, and Hung Truong for their contribution.